\documentclass{article}

\usepackage{arxiv}

\usepackage[utf8]{inputenc} 
\usepackage[T1]{fontenc}    
\usepackage{hyperref}       
\usepackage{url}            
\usepackage{booktabs}       
\usepackage{amsfonts}       
\usepackage{amsmath}
\usepackage{amssymb}
\usepackage{nicefrac}       
\usepackage{microtype}      
\usepackage{lipsum}		
\usepackage{graphicx}
\graphicspath{{gfx/}}
\usepackage{doi}
\usepackage{caption}
\usepackage{subcaption}
\usepackage{float}
\usepackage{svg}
\usepackage{booktabs} 
\usepackage{xcolor}
\usepackage{hyperref}
\usepackage{algpseudocode}
\usepackage{algorithm}
\usepackage{gensymb}
\usepackage{multirow}

\usepackage{multicol}
\newtheorem{lemma}{Lemma}

\usepackage{xspace}
\usepackage{soul}
\usepackage{mathtools}
\usepackage{makecell}
\usepackage{todonotes}
\usepackage{nicefrac}

\newcommand{\ie}{\unskip, i.\,e.,\xspace}
\newcommand{\eg}{\unskip, e.\,g.,\xspace}

\newcommand{\N}{\ensuremath{\mathbb{N}}}

\DeclareMathOperator*{\argmin}{arg\,min}

\definecolor{dgreen}{rgb}{0.0, 0.5, 0.0}

\title{An experimental study of two predictive reinforcement learning methods and comparison with model-predictive control}

\author{ \href{https://orcid.org/0000-0002-1091-7459}{\includegraphics[scale=0.06]{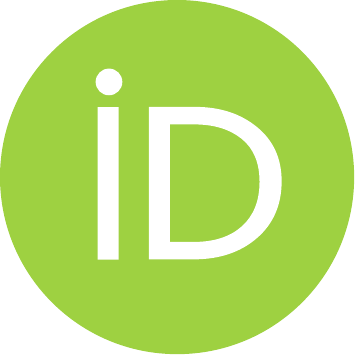}\hspace{1mm}Dmitrii~ Dobriborsci}\thanks{Corresponding author} \\
	Computational and Data Science and Engineering\\
	Skolkovo Institute of Science and Technology\\
	Moscow, Russia \\
	\texttt{d.dobriborsci@skoltech.ru} \\
	\And
	\href{https://orcid.org/0000-0002-6184-3293}{\includegraphics[scale=0.06]{orcid.pdf}\hspace{1mm}Pavel~Osinenko} \\
	Computational and Data Science and Engineering\\
	Skolkovo Institute of Science and Technology\\
	Moscow, Russia \\
	\texttt{p.osinenko@skoltech.ru} \\
}

\renewcommand{\shorttitle}{\textit{Exp. study of two pred. RL methods vs. MPC}}

\hypersetup{
	pdftitle={A template for the arxiv style},
	pdfsubject={q-bio.NC, q-bio.QM},
	pdfauthor={Dmitrii~Dobriborsci},
	pdfkeywords={First keyword, Second keyword, More},
}

\begin{document}
\maketitle
	
\begin{abstract}
Reinforcement learning (RL) has been successfully used in various simulations and computer games. 
Industry-related applications, such as autonomous mobile robot motion control, are somewhat challenging for RL up to date though.
This paper presents an experimental evaluation of predictive RL controllers for optimal mobile robot motion control.
As a baseline for comparison, model-predictive control (MPC) is used.
Two RL methods are tested: a roll-out Q-learning, which may be considered as MPC with terminal cost being a Q-function approximation, and a so-called stacked Q-learning, which in turn is like MPC with the running cost substituted for a Q-function approximation.
The experimental foundation is a mobile robot with a differential drive (Robotis Turtlebot3).
Experimental results showed that both RL methods beat the baseline in terms of the accumulated cost, whereas the stacked variant performed best.
Provided the series of previous works on stacked Q-learning, this particular study supports the idea that MPC with a running cost adaptation inspired by Q-learning possesses potential of performance boost while retaining the nice properties of MPC.
\end{abstract}


\section{Introduction}
Reinforcement Learning (RL) methods achieved great results in recent decade in multiple tasks and competitions, such as Go (strategy board game), chess, shogi (also known as Japanese chess), and even {StarCraft} {II} and Rubic's Cube \cite{akkaya2019solving,silver16mastering,silver18, vinyals19}.
However, only recently RL started moving from rather intuitive playgrounds to technically sound settings such as robot manipulation \cite{kumar2016optimal, borno2013trajectory, tassa2012synthesis} and mobile robot navigation \cite{surmann2020deep}.

RL is closely related to classical control theory \cite{Bertsekas05} which is well-known and widely used in industrial tasks \cite{Soest06, Kouro09, Ma12}. 
In comparison to RL, control algorithms are commonly model-based or at least assume some system properties. 
Model predictive control (MPC) is one of the most recognized algorithms by industry \cite{Qin03, Hrovat12, Darby12, Forbes15, Nikolaou01, Grune12, Mayne14}. 
RL can in turn be divided into two categories: model-free (see \eg \cite{degris2012model, gullapalli1994acquiring, davari2017learning}) and model-based (see \eg \cite{szita2010model, huang2020continual, lambert2020learning}). 
In the former case, actions are optimized based directly on the reward (or running cost) signal.
In the latter, a model of the environment is used to predict future states and rewards (roll-outs). 
Some general advantages and weaknesses of each category are summarized in Table~\ref{table:mb-mf}). 

\begin{table*}[t]
\caption{Model-based vs. model-free reinforcement learning}
\label{table:mb-mf}
\resizebox{\textwidth}{!}{%
\begin{tabular}{@{}|l|l|l|@{}}
\toprule
\multicolumn{1}{|c|}{\textbf{}} & \multicolumn{1}{c|}{\textbf{Model-based RL}}                                                                                                                                                                                         & \multicolumn{1}{c|}{\textbf{Model-free RL}}                                                                                                                                                                        \\ \midrule
Pros                            & \begin{tabular}[c]{@{}l@{}}Can guarantee safety (see \eg \cite{garcia2015comprehensive});\\ \\ Recognized by industry \cite{Polydoros2017survey};\end{tabular} & Plug\&play, easy design, no model needed;                                                                                                                                                                     \\ \midrule
Cons                            & \begin{tabular}[c]{@{}l@{}}Needs accurate environment model;\\ \\ In predictive settings, model error grows with horizon \cite{xiao2019learning};\end{tabular}                                             & \begin{tabular}[c]{@{}l@{}}Industry does not recognize controllers \\ without guarantees \cite{lewis2009reinforcement};\\ \\ Data collection may be expensive \cite{sun2019modelbased};
\end{tabular} \\ \bottomrule
\end{tabular}%
}
\end{table*}

Speaking of robotics, RL appears attractive indeed \cite{Kober2013reinforcement}.
Already in the early 90s, there were attempts at solving leader-following trajectory tracking problem by RL \cite{berns1992reinforcement}.
This could be recognized as a starting point in the application of RL-based control methods for wheeled mobile robots (WMR). 
High dependence of the online learning rate on the previously specified trajectories was noted.
It was recommended to start learning with simple trajectories and then proceed to more complex ones.
Another difficulty was the relatively large number of tuning parameters, which complicates the controller tuning.
The work \cite{beom1995sensor} addressed the problem of autonomous navigation with simultaneous obstacle avoidance using hybrid fuzzy-logic-based RL.
There, the RL-based approach required a supervisor for the initial calibration of the learning model.
It should be noted that only simulation results were presented in these two works.

Group robot control under presence of noise was experimentally tested in \cite{mataric1997reinforcement}.
In recent years, RL-based algorithms took on a new stage of development in robotic applications.
In particular, an optimal path planning problem for a mobile robot was addressed in \cite{low2019solving} based on a variant of Q-learning (QL).
Convergence of QL was boosted by the so-called flower pollination algorithm (FPA).
As MPC is widely recognized as the standard in robotics, RL methods that employ a kind of MPC prediction seem a viable and practicable option.
To these belong \eg roll-out RL schemes \cite{Bertsekas19}, such as that based on QL (abbreviated RQL in the following).

This work investigates RQL in an experimental study with a WMR motion control along with MPC and the so-called stacked QL (SQL) which was suggested in \cite{Osinenko2017stacked} and further provided with stabilizing machinery and generalized to MPC with learning running costs \cite{Beckenbach2018constrained, Beckenbach2018addressing, Beckenbach2019model, Beckenbach2020closed, Beckenbach2020q-learning}.


The main observations can be summarized as follows.
All studied control methods showed better performance at longer prediction horizons, which was expected.
Both RQL and SQL beat the baseline MPC in terms of the accumulated cost.
Still, the stacked variant beat its roll-out counterpart, yet having the same computational complexity.
This shows potential of stacked approaches as viable predictive RL.

The rest of the paper is structured as follows.
After a short introduction, in Sec.~\ref{sec:ps}, a general problem statement, as well as the mathematical model of differential-drive WMR and the control goal are given.
Sec.~\ref{sec:methods} describes studied approaches in the context of the task \ie MPC, RQL, SQL.
The technical description of the experimental platform mobile robot (Robotis TurtleBot3) is presented in Sec.~\ref{sec:setup}.
A summary and analysis of the experimental validation results are given in Sec.~\ref{sec:results}.

\textbf{Notation:}
Sequences: for any $z$: $\{z_{i|k}\}_i^N =$ $\{ z_{1|k}, \dots, z_{N|k} \}$ $= \{ z_k, \dots, z_{k+N-1} \}$, if the starting index $k$ is emphasized; otherwise, it is just $\{z_i\}^N = \{ z_1, \dots, z_N \}$.

\section{Environment and control goal}
\label{sec:ps}

\begin{figure*}[h!]
    \centering
 \begin{subfigure}[b]{0.49\textwidth}
    \centering
        \includegraphics[width=0.75\textwidth]{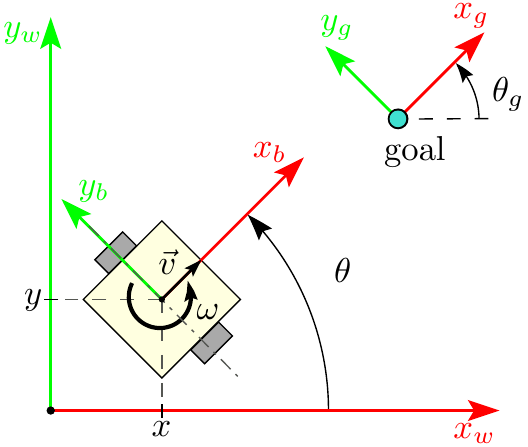}
        \caption{Robot kinematic scheme and a target pose.}
        \label{kinematic}
    \end{subfigure}
    \begin{subfigure}[b]{0.49\textwidth}
        \centering
        \includegraphics[width=0.75\textwidth]{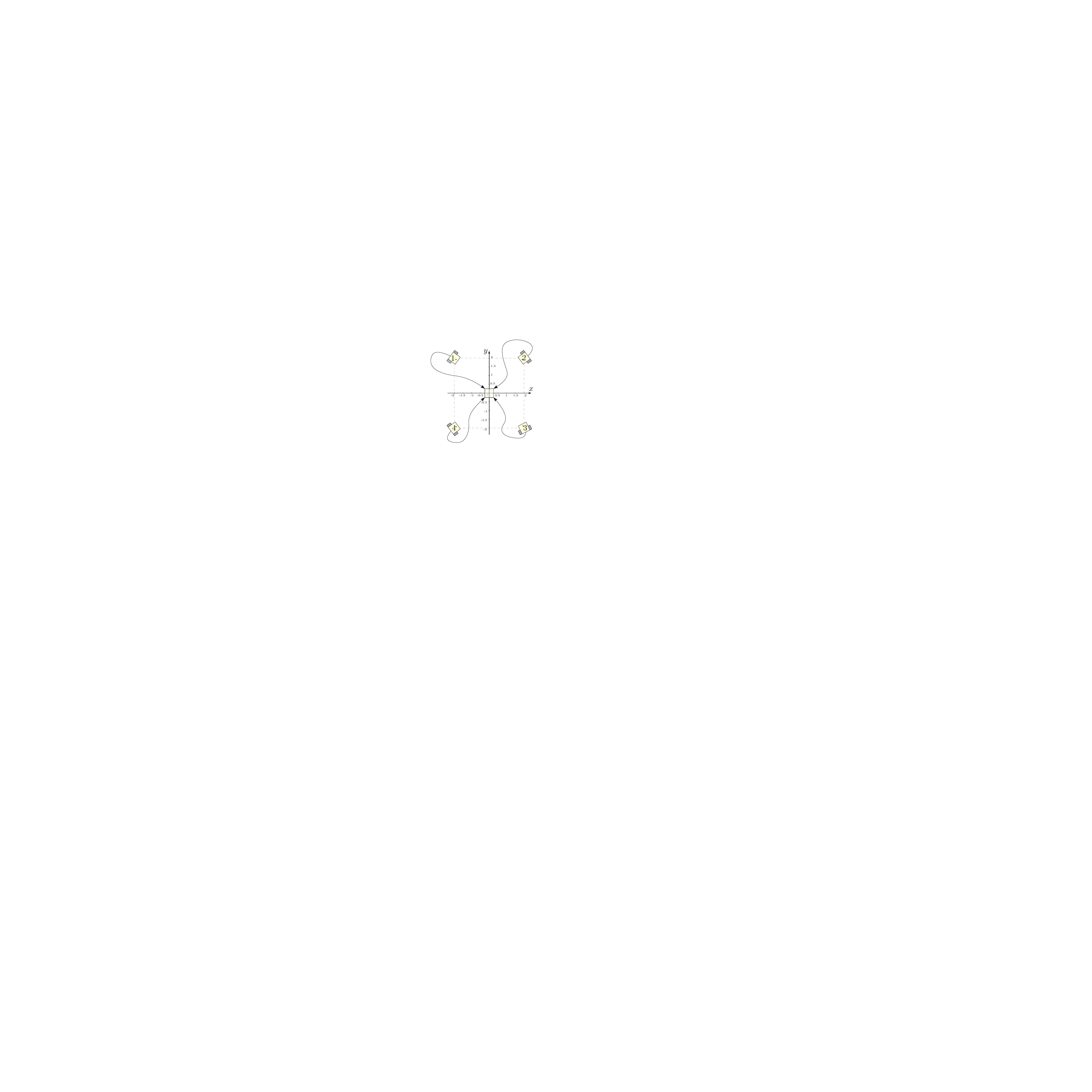}
        \caption{A sketch of starting positions.}
        \label{fig:cg}
    \end{subfigure}
\caption{Experimental system overview}
\label{fig:ctrl-goal}
\end{figure*}

The environment \ie the system, addressed in this study is a differential-drive robot, or diff-drive, which is a particular WMR architecture.
It  consists  of  two  independently driven wheels of radius $R$ that rotate about the same axis, as well as one or more caster wheels, ball casters, or low-friction sliders that keep the robot upright.
Writing the configuration as $q = \big(\begin{matrix}
 x, & y, & \theta
\end{matrix}\big)$, the kinematic equations take the form
\begin{equation}
\label{eqn:3w-robot}
 \begin{cases}
  \dot{x} = v\cos\theta 
   \\
  \dot{y} = v \sin \theta
   \\
   \dot{\theta} = \omega,
 \end{cases}
\end{equation}
where $x$ -- $x$-coordinate [m], $y$ -- $y$-coordinate [m], $\theta$ -- turning angle [rad], $v$ -- velocity [m/s], $\omega$ -- angular velocity [rad/s].
Two advantages of a diff-drive robot are its simplicity (typically the motor is attached directly to the axle of each wheel) and high maneuverability (the robot can spin in place by rotating the wheels in opposite directions). Casters are often not appropriate for outdoor use, however.
The control goal is to stabilize the robot into a desired pose \ie desired coordinates and orientation, from different starting positions (see Fig. \ref{fig:ctrl-goal} for an illustration).
The next section discusses the control methods used.

\section{Methods}
\label{sec:methods}

This section presents the three algorithms used in the experimental study.
It begins with a general optimal control and transitions to MPC followed by RQL and SQL.


\subsection{Model Predictive Control}

A common optimal control problem with a predictive controller implemented in a digital model \ie when the control actions are held constant within sampling intervals, is to minimize a cost function $J$ as follows:
\begin{equation}
    \label{eqn:opt-ctrl}
    \begin{aligned}
        \min_\kappa J&\left(x_0 |\kappa \right) = \min_\kappa \sum_{i=1}^{H} \gamma^{i-1}\rho \left(\hat x_{i|k},  \kappa(x_{i|k})\right), \\
        \textrm{s.t.}~&\mathcal D^+ x = f(x, u^\delta), x(0) = x_0\\
        & x_{i|k} := x((k+i-1) \delta), k \in \N\\
        & u^\delta(t) \equiv u_{i|k} = \kappa(x_{i|k}), t \in \left[ k \delta, (k + i)\delta \right], \\
        &\hat x_{i+1|k} = \Phi \left( \delta, \hat x_{i|k}, u_{i|k} \right),
    \end{aligned}
\end{equation}
where $x$ is the state, $u$ -- action, $\rho$ -- running cost (also stage cost or, more generally, instantaneous objective), $\kappa$ -- control policy, $H$ -- horizon length, $x_k$ -- state at time-step $k$, $\gamma$ -- discounting factor, $\mathcal{D}^{+}$ -- a (generalized) derivative operator, $\delta>0$ -- digital controller sampling time, $\Phi$ describes a state prediction scheme which can be \eg a numerical integrator, say, via Euler method:
\begin{equation}
    \Phi( \delta, \hat x_{i|k}, u_{i|k}) = \hat x_{i|k} + \delta f(\hat x_{i|k}, u_{i|k}).
\end{equation}


In terms of the horizon, $H$ depends on the context of the problem and can be finite ($ H: = N$) or infinite ($H: = \infty $), whence care should be taken of prediction inaccuracy accumulation (a more suitable system description could be pure time-discrete where $\Phi$ produces future state exactly).
When $H: = \infty $, $J$ is also called ``cost-to-go''.
Based on the described variants, two main optimal control formalisms are commonly known, namely Euler-Lagrange and Hamilton-Jacobi-Bellman \cite{Primbs1999-opt-ctrl-CLF}.
The Euler-Lagrange formalism is the foundation of MPC and describes the case when $H$ is finite.
The resulting policy depends on the current state.
The Hamilton-Jacobi-Bellman, in contrast, is used to describe globally optimal policies which only depend on the initial state.
In turn, an infinite horizon can be interpreted as an open horizon -- a situation in which the user is not sure of the exact specification of the horizon. 
The optimal control problem with a finite horizon can be interpreted as an approximation to the problem with an infinite horizon, whereas the latter may well appear intractable.
MPC, in one of its simplest variants, is precisely \eqref{eqn:opt-ctrl} with a finite horizon.
Additional measures, such as terminal costs and constraints, can be used to guarantee closed-loop stability of MPC \cite{Mayne14}.
These can be integrated into RQL and SQL in a similar manner \cite{Gohrt2020reinforcement, Beckenbach2020closed}.
In this study, we concentrate on the performance aspect though, whence stabilizing constraints are omitted.
As for sub-optimality, increasing the horizon reduces the mismatch between the factual cost-to-go under MPC and the value function $V=\min_\kappa J^\kappa$ (the optimized cost-to-go) \cite{Grune08}.
It should be noted that an approximation of the value function is sometimes included into MPC as a terminal cost (cf. RQL).
The idea of SQL in turn is to substitute the running cost for such an approximation altogether.

In general, unlike MPC, RL uses the HJB formalism as the groundwork.
It seeks to approximate the value function not be ``cutting'' the horizon, but by trying to find a solution to the HJB equation (usually via the so-called temporal difference, TD).
While starting with different formalisms, integration of predictive, ``MPC-esque'' elements into RL may actually be viable.
The next two subsections present the respective two RL algorithms.

\subsection{Rollout Q-learning}

A basic actor-critic, value-iteration, on-policy QL reads: 
\begin{equation}
    \label{eqn:QL}
    \begin{array}{lll}
        u_k & := & \argmin \limits_u \hat Q(x_k, u; \vartheta_k), \\
        \vartheta_k & := & \argmin \limits_\vartheta \frac 1 2 \big( \hat Q(x_k, u_k; \vartheta) - \\
        & & \hat Q(x_{k-1}, u_k; \vartheta^-) - \rho(x_k, u_k) \big)^2, \\
    \end{array}
\end{equation}
where $\vartheta$ is vector of the critic neural network weights to be optimized, $\vartheta^-$ is the vector of the weights from the previous time step $k-1$, $\hat Q(\bullet, \bullet; \vartheta)$ -- Q-function approximation parameterized by $\vartheta$.
The latter approximation is effectively done via TD in the value iteration form.
It may be generalized to a custom size experience replay.
Let $e_{j}(\vartheta) := \vartheta \varphi( x_{j-1}, u_{j-1} ) - \vartheta^- \varphi(x_{j}, u_{j}) - \rho(x_{j-1}, u_{j-1}), j \in \N$ denote the TD at time step $k$.
Then, a more general critic cost function may be formulated as
\begin{equation}
	\label{eq:critic-cost}
	J^{c}_k(\vartheta) = \dfrac{1}{2} \sum_{i = k}^{k+M-1} e_{i}^{2}(\vartheta),
\end{equation}
where $M$ is the experience replay \ie buffer, size.
The actor part of RQL reads:
\begin{equation}
    \begin{aligned}
        \min_{\{u_{i|k}\}_{i}^{N}} & J^a_{\text{RQL}} \left( x_k|\{u_{i|k}\}_{i}^{N}; \vartheta_k \right) := \\ 
        & \sum_{i=1}^{N-1} \rho(\hat x_{i|k}, u_{i|k}) + \hat{Q}(\hat x_{N|k}, u_{N|k}; \vartheta_k), \\
        \textrm{s.t.} \quad & \hat x_{i+1|k} = \Phi(\delta, \hat x_{i|k}, u_{i|k}). \\
    \end{aligned}
\end{equation}
It bears a simple form of an action sequence, but a neural network, just like the critic, may be employed, leading to a parameterized actor.
Notice RQL with $N=1$ amounts to the simple data-driven QL \eqref{eqn:QL} which is model-free.
A summary of a TD-based realization of RQL is given in Algorithm \ref{alg:RQL}.

\begin{algorithm*}[h]
    \caption{RQL}
    \label{alg:RQL}
    \begin{algorithmic}
       \State {\bfseries Input:} hyper-parameters (environment model, horizon length $H$, sampling time $\delta$ etc.) 
       \While{True}
       \State Receive system state $x_k$
       \State Critic update: $ \vartheta_k^* := \argmin \limits_{\vartheta} J^{c} = \dfrac{1}{2} \sum \limits_{k = 1}^{N_{c}} e_i^2(\vartheta)$ (see \eqref{eq:critic-cost})
       \State Actor update: $ \{u^*_{i|k}\}^{N}_{i} = \argmin \limits_{\{u_{i|k}\}^{N}_{i}} J^a_{RQL} \left( x_k|\{u_{i|k}\}_{i}^{N}; \vartheta_k^* \right)  = \sum \limits_{i=1}^{N-1} \rho(\hat x_{i|k}, u_{i|k}) + \hat Q(\hat x_{N|k}, u_{N|k}; \vartheta_k^*) $
       \State Apply the first action $u^*_{1|k}$ to the system
       \State Hold the action for $\delta$ seconds
       \EndWhile
    \end{algorithmic}
\end{algorithm*}


\subsection{Stacked Q-learning}

In SQL \cite{Osinenko2017stacked, Beckenbach18, Beckenbach2018constrained, Beckenbach2018addressing, Beckenbach2019model, Beckenbach2020closed, Beckenbach2020q-learning}, a single Q-function approximation is extended to a sum, which results in a finite stack of Q-function approximations over a finite number of time steps (see Algorithm~\ref{alg:SQL}).
In turn, approximation of Q-function may be sought not only for the current state, but also for the series of future Q-functions.
Let us provide some groundwork for SQL.
The generic Q-function under a policy $\kappa$ reads, for a state $x_k$:
\begin{equation}
    Q^{\kappa}(x_k, u_k) = \rho(x_k, u_k) + J^{\kappa}(x_{k+1}).
\end{equation}
The value function $V$ and the Q-function are related as follows:
\begin{equation}
    \begin{aligned}
        & V(x_k) := \min_{u_k} Q(x_k, u_k) = \rho(x_k, u_k) + V(x_{k + 1}).
    \end{aligned}
\end{equation}
A Q-function stack can be defined as \cite{Osinenko2016-stacked-ADP}:
\begin{equation}
    \begin{aligned}
        \bar{Q} \left(x_k,  \{ u_{i|k}\}_i^{N} \right) := \sum_{i=1}^{N} Q(x_{i|k}, u_{i|k}) \\
    \end{aligned}.
\end{equation}
Under generic policies, the stack can be expressed as:
\begin{equation}
    \begin{aligned}
        {\bar{Q}}^{\{\kappa_{i|k}\}_i^{N}} (x_k,  \{ u_{i|k}\}_i^{N}) & = \sum_{i=1}^{N}  \rho \left(x_{i|j}, u_{i|k}\right) + \\ 
        & + \sum_{i=1}^{N} \sum_{j=k+1}^{\infty}  \rho \left(x_{i|j},\kappa_{i}(x_{i|j})\right),
        \label{eq:stacked-value-function}
    \end{aligned}
\end{equation}
where $i$ is a horizon index and $j$ is an index for starting state $k$ update, $\{\kappa_{i|1}\}_i^{N}$ -- stack of policies.
Compare it to a single Q-function case:
\begin{equation}
    Q^{\kappa}(x_{k}, u_k) = \rho \left(x_k, u_k \right) + \sum_{j=k+1}^{\infty} \rho \left(x_{j},\kappa (x_j) \right).
    \label{eq:value-function}
\end{equation}
The next lemma shows that SQL is a valid variant of dynamic programming.
Namely, the optimal policy of the ordinary QL and, respectively, SQL yield the same cost.
\begin{lemma}[\cite{Beckenbach18}]
For the time step $k$, let Q-function $Q(x,u)$ be defined by 
\eqref{eq:value-function}
and the stacked Q-function $\bar{Q}(x,u)$ by
\eqref{eq:stacked-value-function}.
Then
\begin{align}
    \bar{Q} \left(x_k, \{ u_{i|k}\}_{N} \right) =  \sum_{i=1}^{N} Q(x_{i|k}, u_{i|k}) , \quad \forall k \in \N_{\geq 0},				
    \label{lemma:V*-leq-Gamma}
\end{align}
where $Q$ is the single optimal Q-function and $\bar{Q}$ is the stacked optimal Q-function. 
\end{lemma}
Practically, SQL can be realized via a critic minimizing the TD in a similar manner as in RQL. 
The SQL actor in turn reads:
\begin{equation}
    \begin{aligned}
        \min_{\{u_{i|k}\}_{i}^{N}} \quad & J^a_{\text{SQL}} \left( x_k|\{u_{i|k}\}_{i}^{N}; \vartheta_k \right) = \sum_{i=1}^{N} \hat{Q}(\hat x_{i|k}, u_{i|k}; \vartheta_k), \\
        \textrm{s.t.} \quad & \hat x_{i+1|k} = \Phi(s \delta, \hat x_{i|k}, u_{i|k}).
    \end{aligned}
\end{equation}
Having described the control methods, we proceed to the next section that describes the experimental part of this work.
\begin{algorithm*}
	\caption{SQL}
	\label{alg:SQL}
	\begin{algorithmic}    
		\State {\bfseries Input:}hyper-parameters (environment model, horizon length $H$, sampling time $\delta$ etc.)  
		\While{True}
		\State Receive system state $x_k$
		\State Critic update: $\vartheta_k^* := \argmin \limits_{\vartheta_k} J^c = \dfrac{1}{2} \sum \limits_{k = 1}^{N_{c}} e^2_{k}(\vartheta_k)$ See eq. \eqref{eq:critic-cost}
		\State Actor update: $\{u^*_{i|k}\}^{N}_{i} = \argmin \limits_{\{u_{i|k}\}^{N}_{i}} J^a_{SQL} \left( x_k|\{u_{i|k}\}_i^{N}; \vartheta_k^* \right) := \sum \limits_{i=1}^{N} \hat Q(\hat x_{i|k}, u_{i|k}; \vartheta^*_k)$
		\State Apply the first action $u^*_{1|k}$ to the system
		\State Hold the action for $\delta$ seconds
		\EndWhile
	\end{algorithmic}
\end{algorithm*}

\section{Experimental setup}
\label{sec:setup}
\begin{figure*}[h!]
    \centering
    \begin{subfigure}[b]{0.49\textwidth}
        \includegraphics[width=\textwidth]{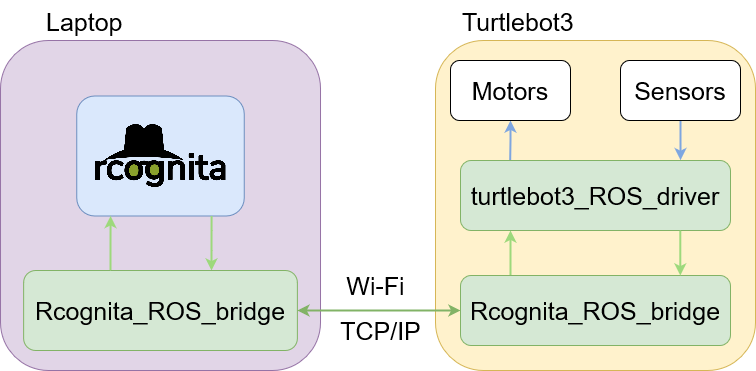}
        \caption{Software diagram of the experimental setup.}
    \end{subfigure}
    \begin{subfigure}[b]{0.49\textwidth}
        \includegraphics[width=\textwidth]{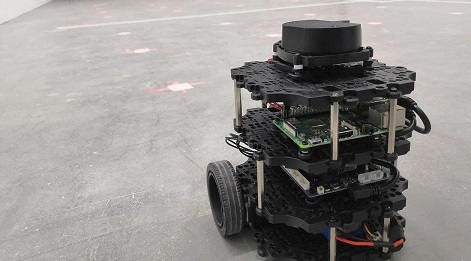}
        \caption{Robotis TurtleBot3}
    \end{subfigure}
\caption{Overview of the experimental setup.}
\label{fig:setup}
\end{figure*}

Implementation of MPC, RQL, and SQL was done using a custom Python package, developed specifically for hybrid simulation of RL agents called \texttt{rcognita} \footnote{\url{https://github.com/AIDynamicAction/rcognita}}.
This package was extended to account for integration with the Robot Operating System (ROS), which is a common tool for rapid prototyping in robotics.

\subsection{Robotis Turtlebot3}


The hardware platform in the experimental study was a Robotis Turtlebot 3 equipped with a lidar, a board computer, power electronics and an inertial measurement unit (IMU).
All the components are compatible with ROS, which was essential in controlling the robot in a convenient way.
The lidar, which is a 360$^\circ$ laser scanner, along with the IMU allowed for a precise localization.
The robot itself has a maximal linear velocity of 0.22 m/s, an angular velocity of 2.48 rad/s and can carry a payload of up to 15 kg.

\subsection{Method evaluation details}

The running cost was considered in the following quadratic form:
\begin{equation}
    \begin{aligned}
        & \rho = \chi^\top R \chi, \\
    \end{aligned}
\end{equation}
where $\chi = [x, u]$, $R$ diagonal, positive-definite.
The critic structure was chosen quadratic as follows:
\begin{equation}
    \begin{aligned}
        & \hat{Q}(x,u; \vartheta) := \vartheta \varphi^\top (x, u), \\
        & \varphi (x, u) := \textrm{vec} \left(\Delta_{u} \left([x|u] \otimes [x|u] \right) \right), \\
    \end{aligned}
\end{equation}
where $\vartheta$ -- critic weights, $\varphi$ -- critic activation function,
$\Delta_{u}$ -- operator of taking the upper triangular matrix, $\textrm{vec}$ -- vector-to-matrix transformation operation, $[x|u]$ -- stack of vectors $x$ and $u$,  $\otimes$ -- Kronecker product. 
The prediction of the robot's states was carried out using the model \eqref{eqn:3w-robot}.
For the evaluation of the control methods, a number of experimental runs for different starting positions were carried out (see Fig.~\ref{fig:cg}).
For each starting position, five runs were performed for each control method.
Initially, the position of the robot $p_b^w = \begin{bmatrix}
x & y & \theta \end{bmatrix}$  and the goal $p_g^w = \begin{bmatrix}
x_g & y_g & \theta_g \end{bmatrix}$ were given in the same frame (see Fig.~\ref{kinematic}). 
The coordinates of the robot were then expressed in the target frame \cite{spong2020robot}.
At this step, we used the transformation matrix between the robot's frame and the target's frame in the following form:
\begin{equation*}
    T(\theta_g) = \begin{bmatrix}
\cos\theta_g & -\sin\theta_g & 0 & x_g\\ 
\sin\theta_g & \cos\theta_g & 0 & y_g\\
0 & 0 & 1 & 0 \\
0 & 0 & 0 & 1
\end{bmatrix}.
\end{equation*}
Therefore, the frame transformation took the form:
\begin{equation*}
    p_g^b = T p_b^w.
\end{equation*}

\begin{figure}
    \centering
    \begin{subfigure}[t]{0.49\textwidth}
        \includegraphics[width=\linewidth]{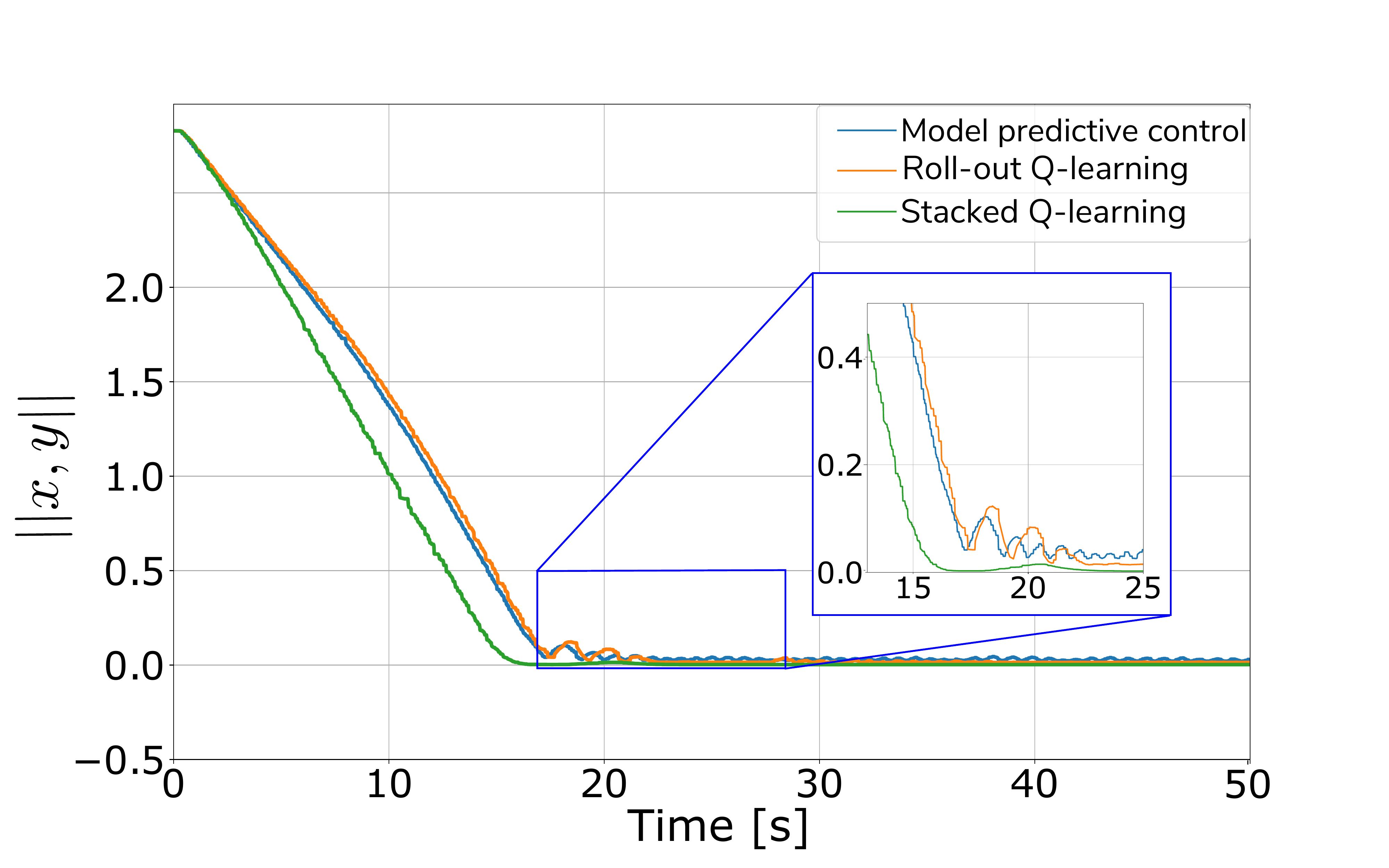}
        \caption{Distance to the goal.}
    \end{subfigure}
    \begin{subfigure}[t]{0.49\textwidth}
        \includegraphics[width=\linewidth]{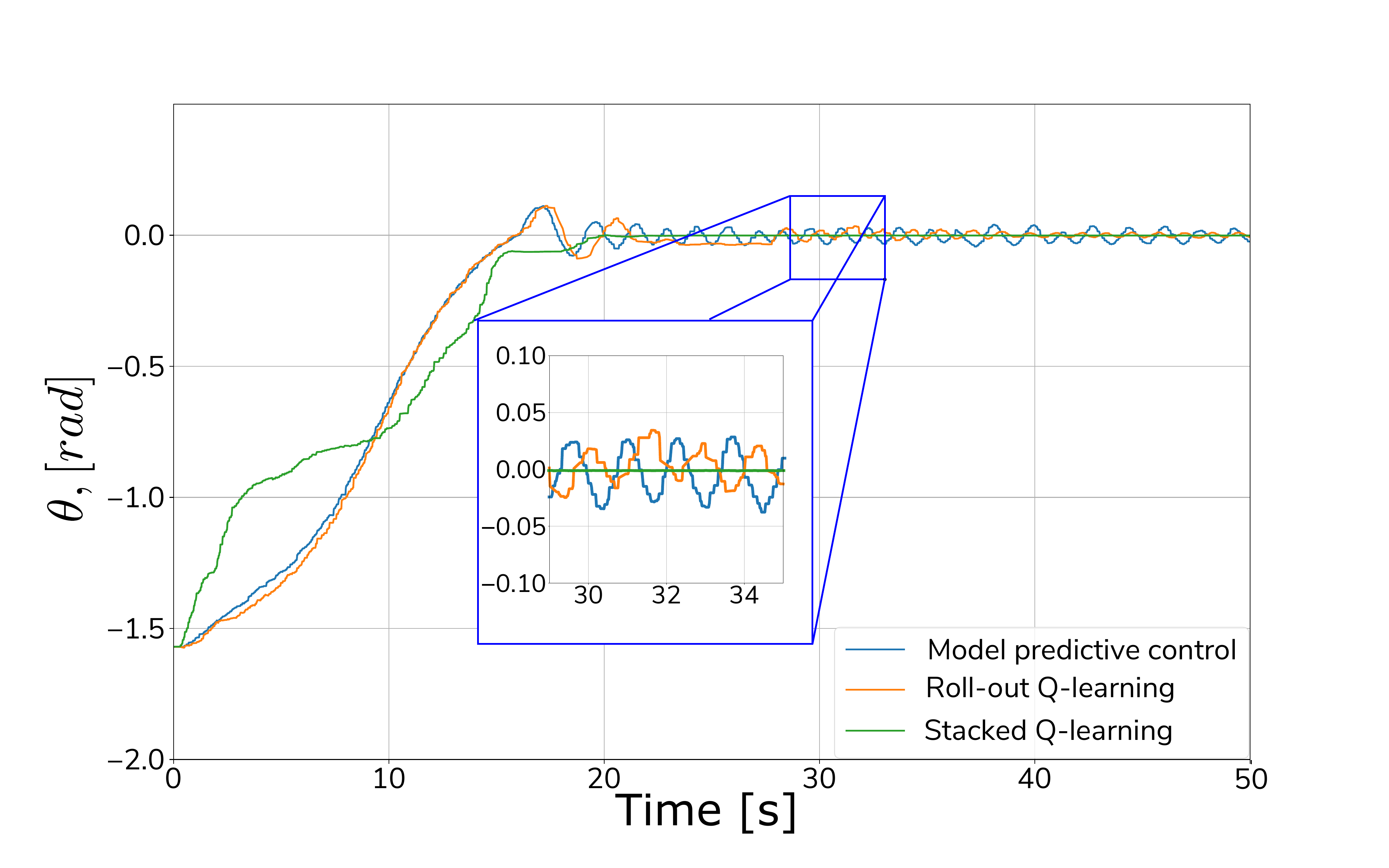}
        \caption{Orientation of the robot.}
    \end{subfigure}
    \caption{Transients for the starting point 1 at long horizon (N = 1.2 seconds).}
    \label{fig:trans-long}
\end{figure}

\begin{figure}
    \centering
    \begin{subfigure}[t]{0.49\textwidth}
        \includegraphics[width=\linewidth]{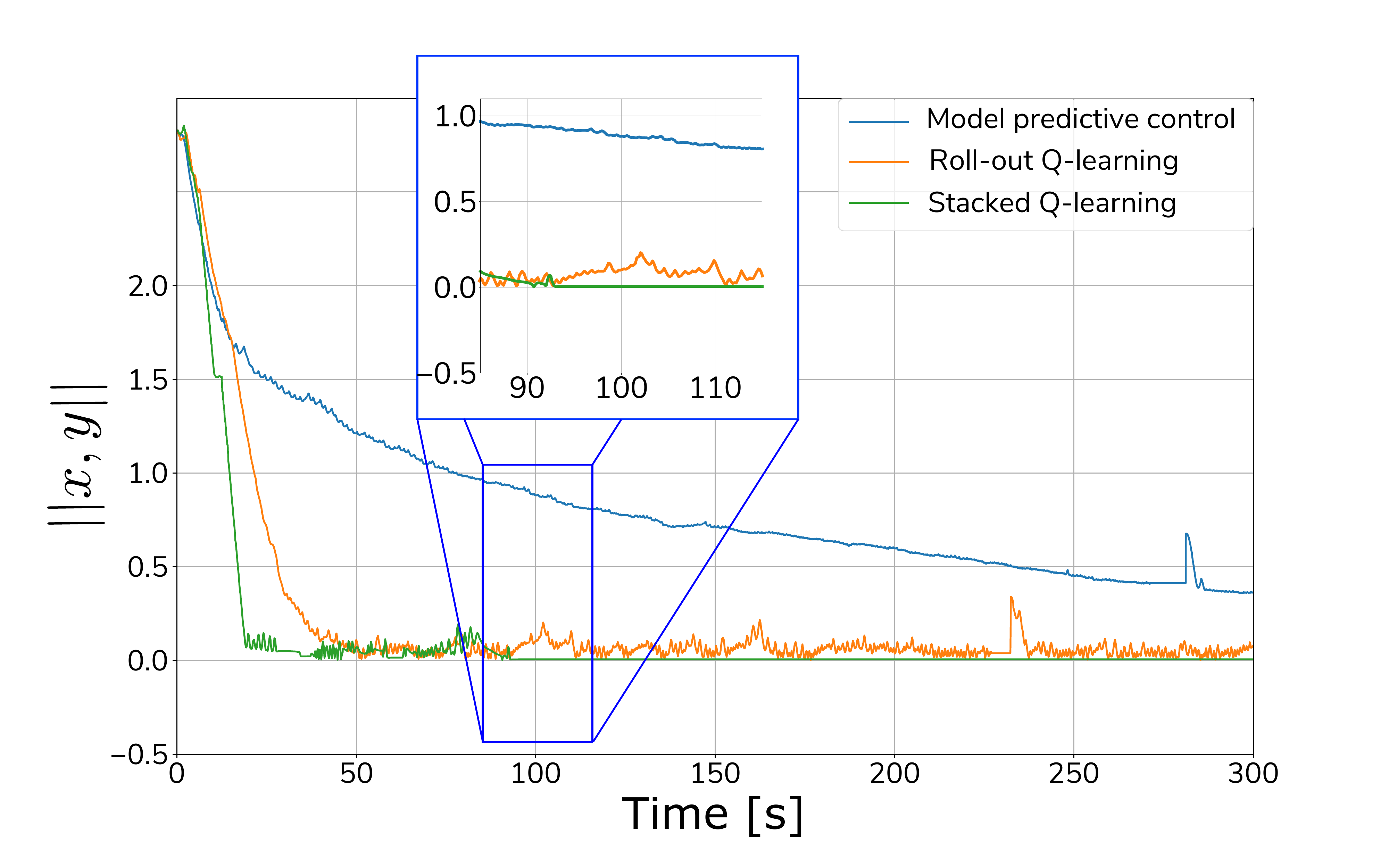}
        \caption{Distance to the goal.}
    \end{subfigure}
    \begin{subfigure}[t]{0.49\textwidth}
        \includegraphics[width=\linewidth]{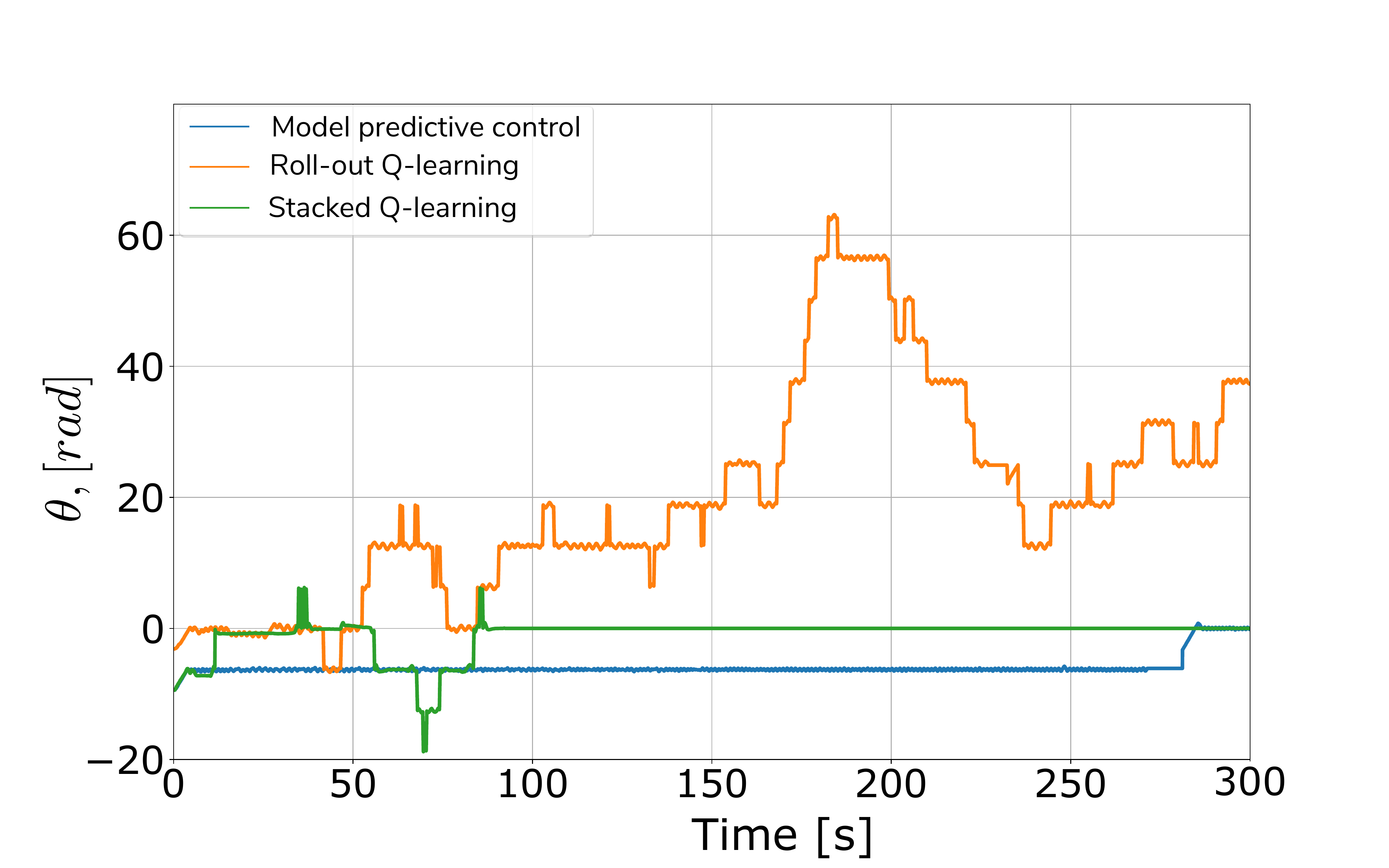}
        \caption{Orientation of the robot.}
    \end{subfigure}
    \caption{Transients for the starting point 1 at short horizon (0.2 seconds).}
    \label{fig:trans-short}
\end{figure}

\begin{figure}
\centering
    \begin{subfigure}[t]{0.49\textwidth}
        \includegraphics[width=\linewidth]{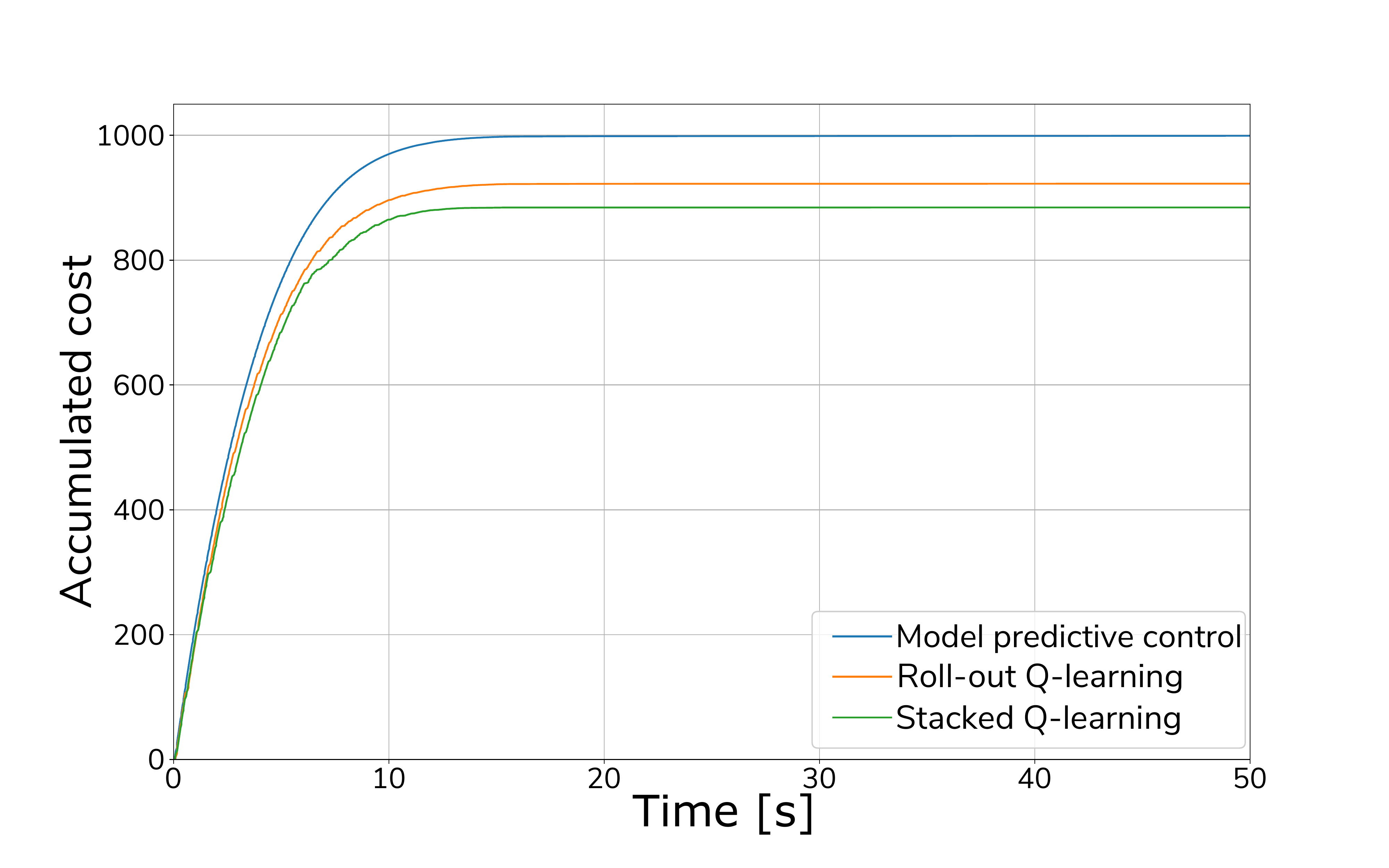}
        \caption{Accumulated cost at long horizon (1.2 sec).}
        \label{fig:long-cost}
    \end{subfigure}
    \begin{subfigure}[t]{0.49\textwidth}
        \includegraphics[width=\linewidth]{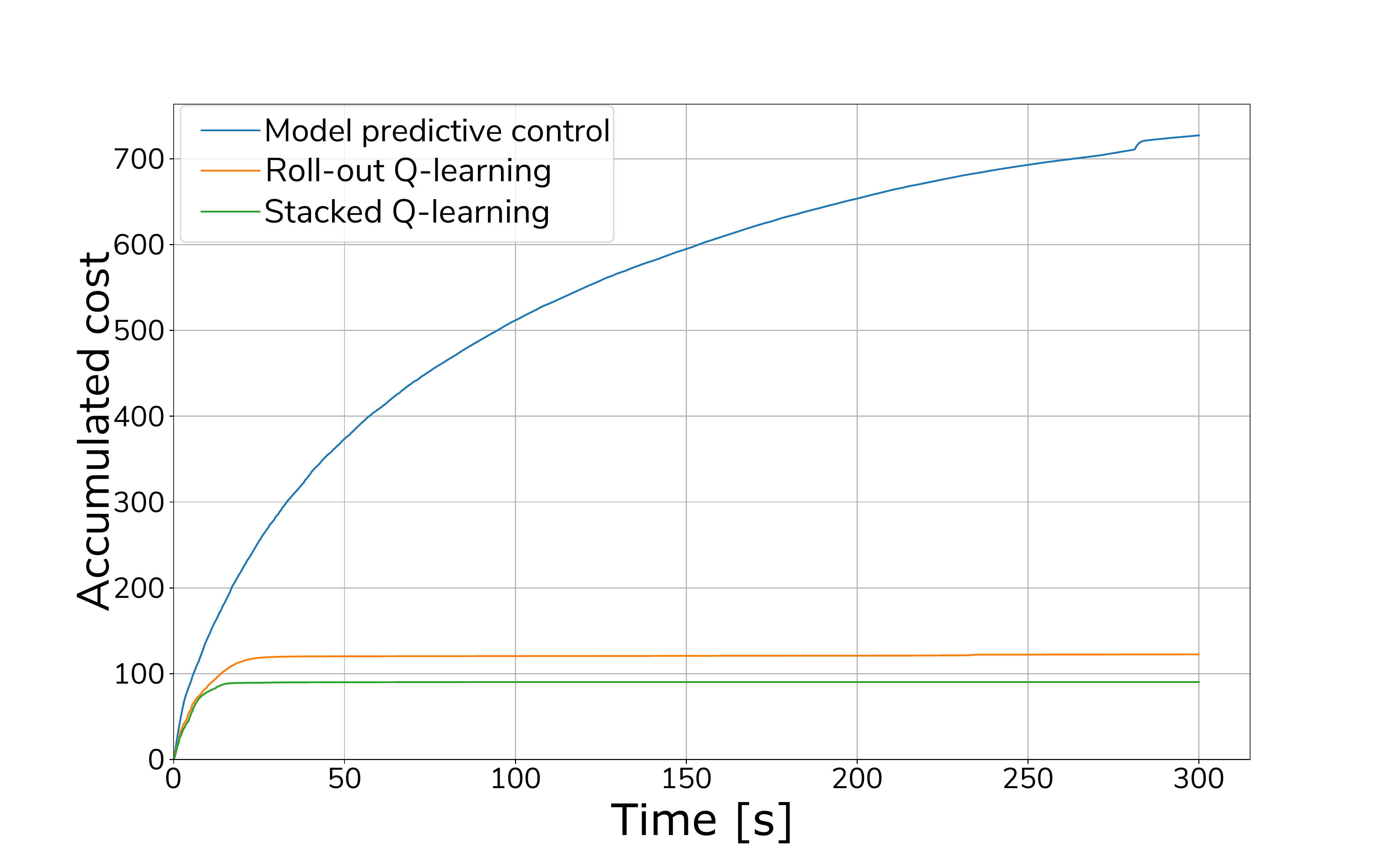}
        \caption{Accumulated cost at short horizon (0.2 sec).}
    \end{subfigure}
    \caption{Cost comparison.}
    \label{fig:cost-traj}
\end{figure}

\begin{figure}
\centering
    \begin{subfigure}[t]{0.49\textwidth}
        \includegraphics[width=\linewidth]{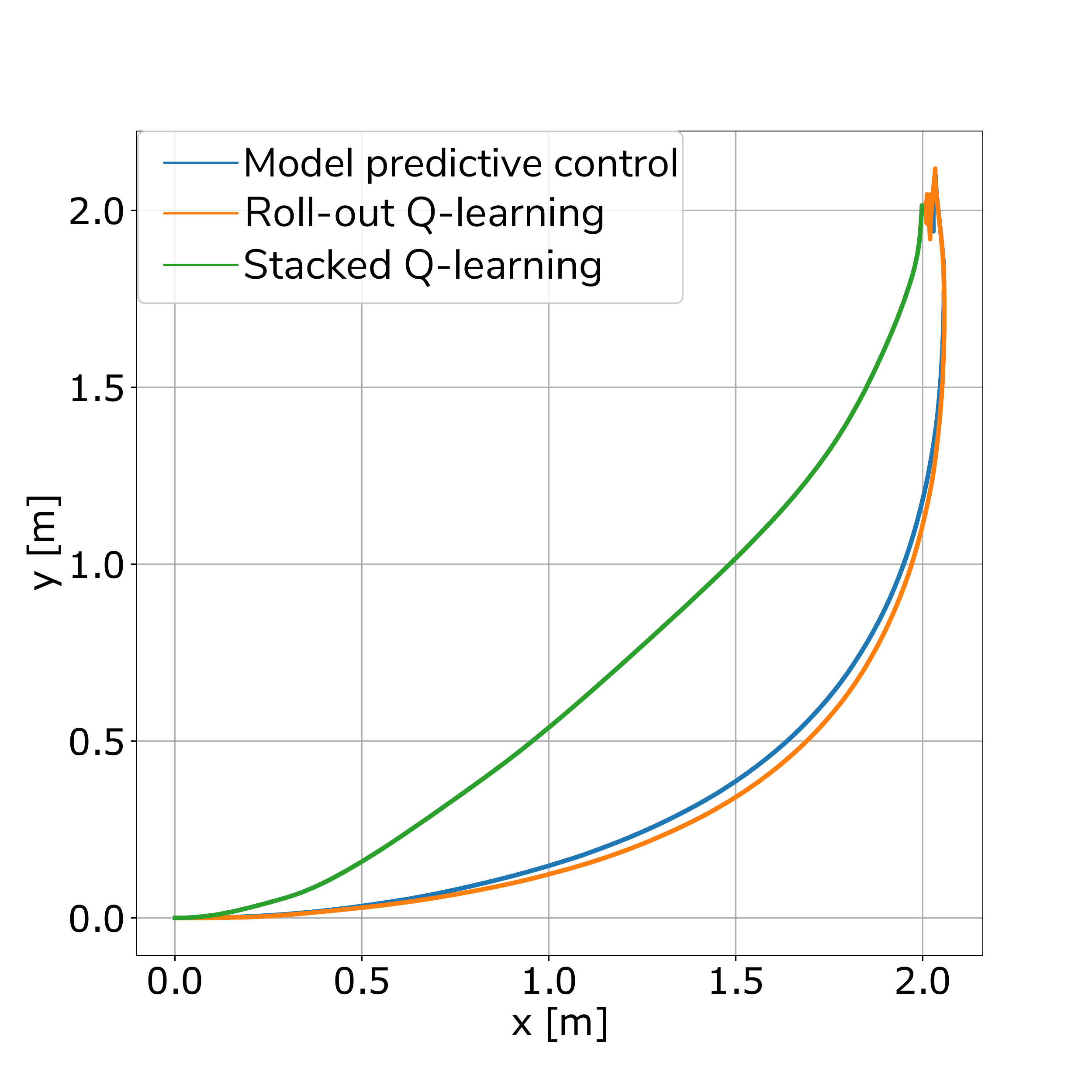}
        \caption{Robot trajectories at long horizon (1.2 sec).}
        \label{fig:short-cost}
    \end{subfigure}
    \begin{subfigure}[t]{0.49\textwidth}
        \includegraphics[width=\linewidth]{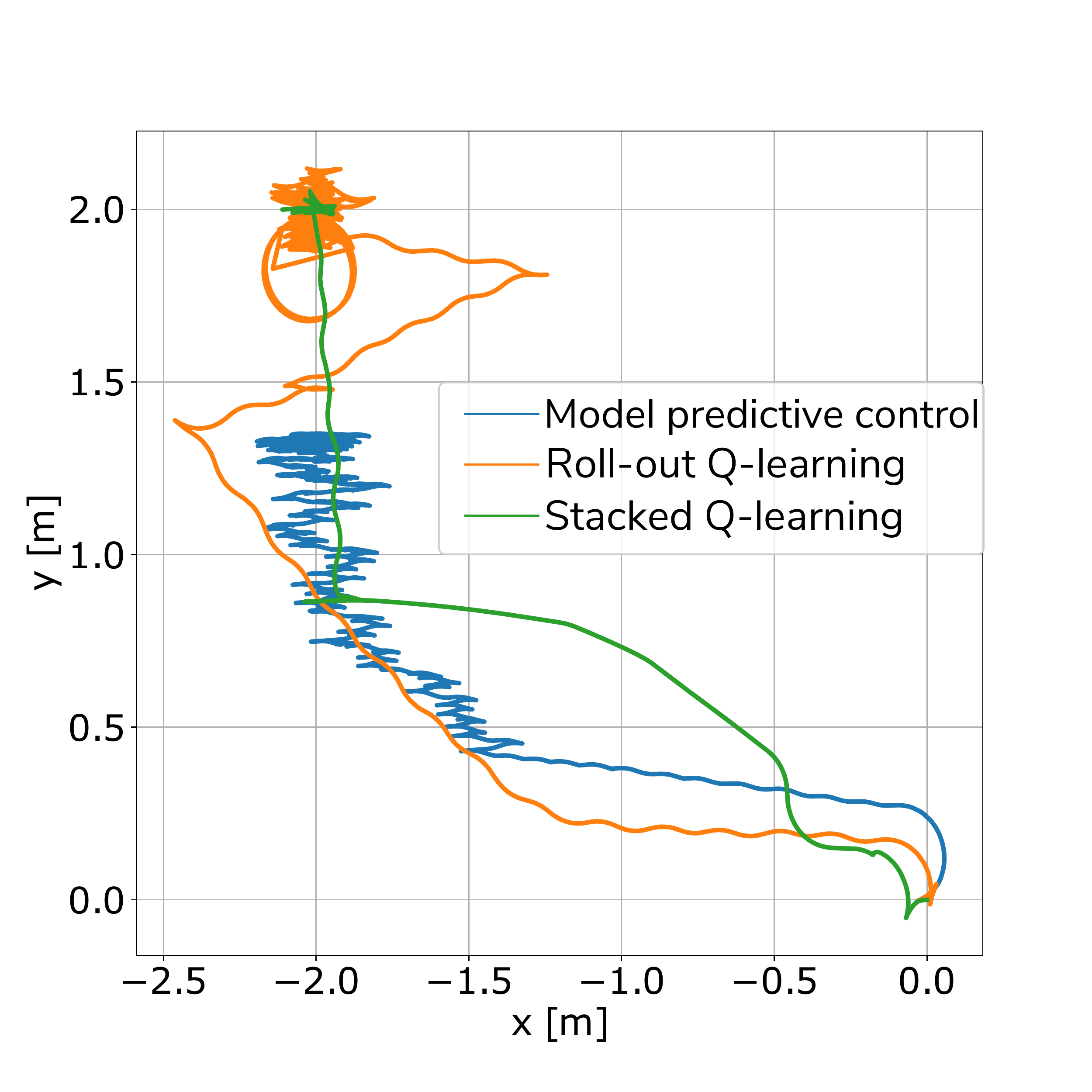}
        \caption{Robot trajectories at short horizon (0.2 sec).}
        \label{fig:short-traj}
    \end{subfigure}
    \caption{Experimental results: robot trajectories under short- and long-horizon control methods.}
    \label{fig:cost-traj-short}
\end{figure}

\section{Results and discussion}
\label{sec:results}
As can be seen from the results of experimental evaluation (Fig.~\ref{fig:trans-long}~--~\ref{fig:cost-traj-short}), all three algorithms were close in convergence rates in the case of long prediction horizons.
However, SQL provided higher accuracy and lower energy cost (Fig.~\ref{fig:long-cost}).

Both the RQL and SQL generally outperformed MPC, especially at short horizons (see fig.~\ref{fig:trans-short}~--~\ref{fig:cost-traj-short}), where (supposedly) learning-based elements dominated.
An interesting observation was, however, a clear dominance of SQL in terms of stabilization time and energy cost.
Fig.~\ref{fig:short-cost} shows the average value of the accumulated cost for each algorithm launches at different starting positions.
It should be noted that RQL and SQL differ from MPC in terms of complexity only by the critic part, which is the same for both.
Thus, RQL and SQL are of the same complexity class.
Yet, considering the performance boost of SQL observed in the experiments, along with the previous results \cite{Osinenko2017stacked, Beckenbach2018constrained, Beckenbach2018addressing, Beckenbach2019model, Beckenbach2020closed, Beckenbach2020q-learning}, further supports the idea that predictive RL, that base on running cost adaptation inspired by the HJB formalism, may be of interest in robotics.
They retain the positive features of MPC yet offer potential of performance improvement.


\bibliographystyle{model1-num-names}


\end{document}